\pdfoutput=1

\documentclass[11pt]{article}

\usepackage[final]{acl}

\usepackage{times}
\usepackage{latexsym}

\usepackage[T1]{fontenc}

\usepackage[utf8]{inputenc}

\usepackage{microtype}

\usepackage{inconsolata}

\usepackage{booktabs}
\usepackage{multirow}
\usepackage{multicol}
\usepackage{makecell}
\usepackage{enumitem}
\usepackage{balance}
\usepackage{graphicx}
\usepackage{subfigure}
\usepackage{amsmath}
\usepackage{amsthm}
\usepackage{amssymb}
\usepackage{amsfonts}
\usepackage{hyperref}
\usepackage{enumitem}
\usepackage{xparse}
\usepackage[most]{tcolorbox}
\usepackage{xspace}

\usepackage{soul}    
\usepackage{xcolor}  
\usepackage{colortbl}
\usepackage{booktabs}

\definecolor{mintleaf}{RGB}{0, 184, 148}
\definecolor{dm-blue-500}{RGB}{0, 69, 177}
\definecolor{dm-purple-500}{RGB}{105,50,230}
\definecolor{mysilver}{RGB}{128,129,128}
\definecolor{my_green}{RGB}{0, 176, 80}
\definecolor{my_yellow}{RGB}{255,165,0}
\definecolor{my_red}{RGB}{255, 0, 0}
\definecolor{my_purple}{RGB}{126, 100, 158}
\definecolor{my_blue}{RGB}{49, 133, 155}
\definecolor{case_purple}{RGB}{160, 43, 147}
\definecolor{case_blue}{RGB}{15, 158, 213}

\newcommand{\model}{STeCa\xspace}

\definecolor{deepgreen}{RGB}{0,100,0}
\definecolor{deepred}{RGB}{139,0,0}

\title{STeCa: Step-level Trajectory Calibration for LLM Agent Learning}

\author{
Hanlin Wang, ~~Jian Wang$^{\dagger}$, ~~Chak Tou Leong, ~~Wenjie Li \\
Department of Computing, The Hong Kong Polytechnic University \\
\texttt{\{hanlin-henry.wang,chak-tou.leong\}@connect.polyu.hk} \\
\texttt{jian51.wang@polyu.edu.hk} ~~
\texttt{cswjli@comp.polyu.edu.hk}
}

\begin{document}
\maketitle

\renewcommand{\thefootnote}{$\dagger$}
\footnotetext[1]{Corresponding author.}
\setcounter{footnote}{0}
\renewcommand{\thefootnote}{\arabic{footnote}}

\begin{abstract}

Large language model (LLM)-based agents have shown promise in tackling complex tasks by interacting dynamically with the environment. 
Existing work primarily focuses on behavior cloning from expert demonstrations or preference learning through exploratory trajectory sampling. 
However, these methods often struggle to address long-horizon tasks, where suboptimal actions accumulate step by step, causing agents to deviate from correct task trajectories.
To address this, we highlight the importance of \textit{timely calibration} and the need to automatically construct calibration trajectories for training agents. We propose \textbf{S}tep-Level \textbf{T}raj\textbf{e}ctory \textbf{Ca}libration (\textbf{\model}), a novel framework for LLM agent learning. 
Specifically, \model identifies suboptimal actions through a step-level reward comparison during exploration. It constructs calibrated trajectories using LLM-driven reflection, enabling agents to learn from improved decision-making processes. We finally leverage these calibrated trajectories with successful trajectories for reinforced training.
Extensive experiments demonstrate that \model significantly outperforms existing methods. Further analysis highlights that timely calibration enables agents to complete tasks with greater robustness. 
Our code and data are available at \url{https://github.com/WangHanLinHenry/STeCa}.

\end{abstract}

\section{Introduction}

Large language models (LLMs) have recently demonstrated impressive capabilities in reasoning and planning, leading to the development of various LLM-based agents that tackle real-world tasks such as household assistance~\citep{puig2018virtualhome,shridhar2020alfworld}, web browsing~\citep{yao2022webshop,deng2023mind2web}, and complex scientific reasoning~\citep{wang-etal-2022-scienceworld}. 
These agents typically engage in long-horizon interactions with environments, making sequential decisions to accomplish specific goals.
Despite their promise, LLM agents still face significant challenges in generating high-quality task plans for complex scenarios~\citep{xie2024revealing,wang-etal-2024-e2cl}. This highlights the need for more effective approaches to enhance their decision-making capabilities over time.

\begin{figure}[t!]
    \centering
    \includegraphics[width=0.98\linewidth]{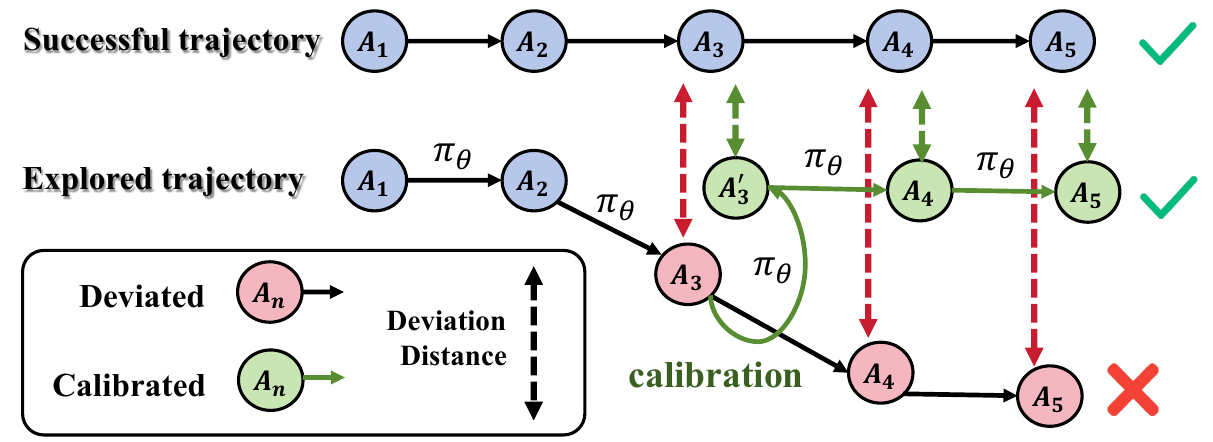}
    \caption{Step-level calibration enables LLM agents to construct calibrated trajectories and learn to mitigate the accumulation of suboptimal actions.}
    \label{fig:intro}
    \vspace{-9pt}
\end{figure}

Previous work has investigated agent learning by leveraging augmented exploratory data ~\citep{chen2023fireact,yin2023lumos,zeng2023agenttuning,xiang2024language}. These methods primarily rely on behavior cloning from expert demonstrations, training agents exclusively on successful trajectories. However, this approach prevents agents from proactively self-correcting mistakes, leading to the accumulation of errors and ultimately suboptimal task performance~\citep{xie2024revealing}.
To address this limitation, another line of work focuses on preference learning~\citep{song2024trial, xiong2024watch} and reinforcement learning~\citep{carta2023grounding,tan2024true}, integrating failure trajectories additionally to refine decision-making. These approaches train LLM agents using explicit error signals or reward functions. 
However, many long-horizon agentic tasks involve multi-turn interactions, where errors often only become evident at the terminal state~\citep{yuan2025agent}. As a result, these methods fail to address early-stage deviations, which may not be immediately apparent but accumulate incrementally over time, ultimately leading to significant errors.

To address the above limitations, we highlight the importance of \textbf{timely calibration}, which allows agents to \textit{adjust suboptimal actions as they occur, rather than deferring correction until after the entire exploration}. 
As illustrated in Figure~\ref{fig:intro}, when an early suboptimal action occurs, the subsequent actions are prone to deviate from the optimal trajectory, significantly increasing the risk of task failure. If an agent can engage in self-reflection and calibrate its behavior in real time, it stands a much better chance of successfully completing the task. 
However, implementing step-level calibrations presents significant challenges. 
1) Unlike in mathematical reasoning tasks~\citep{kumar2024training,xi2025rise}, where well-defined rules simplify error detection, identifying deviations at each step in long-horizon agentic tasks is considerably more complex. This complexity stems from the dynamic and diverse nature of task execution in interactive environments. 
2) As far as we know, the lack of step-level calibration trajectory data poses a major obstacle to training agents to effectively recognize and mitigate deviations.

In this work, we propose \textbf{S}tep-level \textbf{T}raj\textbf{e}ctory \textbf{Ca}libration (\textbf{\model}), a novel agent learning framework that enables LLM agents to perform real-time calibration. 
\model operates by interacting with the environment to perform explorations and utilizes Monte Carlo (MC) sampling~\citep{kakade2002approximately} to estimate step reward for each action. 
By comparing the rewards of adjacent actions, \model effectively identifies deviated actions that lead to suboptimal performance.
Then, we utilize off-the-shelf LLMs for reflection, which revises a deviated action into its ground-truth counterpart while generating a reflective thought. The resulting action and its thought, along with subsequent expert trajectory, form a \textit{calibrated trajectory}.
All calibrated trajectories, combined with successful trajectories during exploration, are then used to reinforce the agent's training, optimizing its learning process.
We evaluate \model on two widely-used agent benchmarks~\citep{puig2018virtualhome,shridhar2020alfworld}. Extensive experimental results demonstrate that \model significantly outperforms existing methods, achieving higher success rates across a variety of tasks.

In summary, our contributions are as follows:
\begin{itemize}[leftmargin=*, nolistsep]
\setlength{\itemsep}{1mm}
\item We highlight the importance of timely calibration in interactive agentic tasks, a crucial aspect largely overlooked by previous methods.
Unlike existing approaches that rely on terminal-state error signals or reward functions, we emphasize the need for real-time adjustments to prevent the accumulation of deviations, which can lead to significant errors in long-horizon tasks.
\item We introduce \model, a novel learning framework that enhances LLM agents by integrating an automated deviation detection mechanism and calibrated trajectory construction. It equips agents with essential calibration capabilities for improvement during task execution.
\item Extensive experiments demonstrate that \model significantly outperforms existing methods. By detecting deviations in real-time, \model enables agents to effectively mitigate the accumulation of suboptimal actions and handle long-horizon tasks more robustly.
\end{itemize}

\section{Preliminaries}

\paragraph{Task Formulation.}
\label{sec:task_form}
This work investigates how LLM-based agents tackle long-horizon tasks within specific environments through interactions.
Following previous studies~\citep{song2024trial, xiong2024watch}, we formalize these agentic tasks as a partially observable Markov decision process (POMDP), which contains the key elements ($\mathcal U, \mathcal S, \mathcal A, \mathcal O, \mathcal T, \mathcal R$). Here, $\mathcal U$ denotes the instruction space, $\mathcal S$ the state space, $\mathcal A$ the action space, $\mathcal O$ the observation space, $\mathcal T$ the transition function ($\mathcal T: \mathcal S \times \mathcal A \rightarrow \mathcal S$), and $\mathcal R$ the reward function ($\mathcal R: \mathcal S \times \mathcal A \rightarrow [0, 1]$). Since the task planning capability of LLM agents is our main focus, $\mathcal U, \mathcal A, \mathcal O$ are subsets of natural language space.

Given a task instruction $u\in\mathcal{U}$, the LLM agent $\pi_{\theta}$ at time step $t$ takes an action $a_t\sim \pi_\theta(\cdot|u, e_{t-1})$ and receives the environmental feedback as the observation $o_t\in\mathcal{O}$. $e_{t-1}$ denotes the historical interaction trajectory $(a_1, o_1, ... , a_{t-1}, o_{t-1})$. Each action $a_t$ incurs the environment state to $s_t\in\mathcal{S}$. The interaction loop terminates when either the agent completes the task or the maximum step is reached.
The final trajectory is $e_m = (u, a_1, o_1, ..., a_m, o_m)$, where $m$ denotes the trajectory length. The outcome reward $r_o(u, e_m) \in [0, 1]$ indicates the success or failure of the task.

\begin{figure*}[t!]
    \centering
    \includegraphics[width=0.95\textwidth]{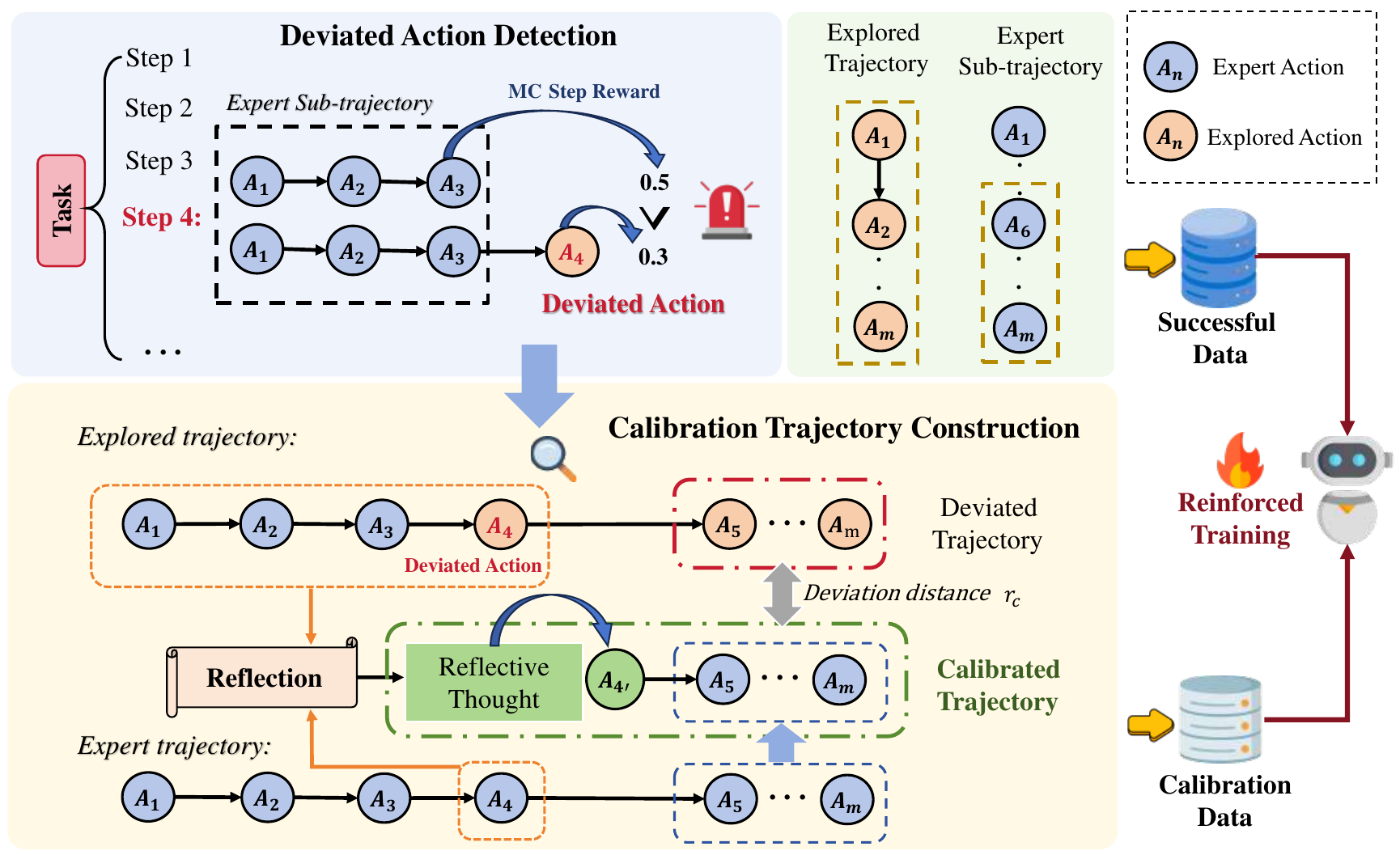}
    \caption{Overview of the \textbf{S}tep-level \textbf{T}raj\textbf{e}ctory \textbf{Ca}libration (\textbf{STeCa}) framework for LLM agent learning.}
    \label{fig:overview}
\end{figure*}

\paragraph{Step-level Reward Acquisition.}
\label{sec:step_reward_compute}

It is crucial to acquire step-level rewards as feedback to improve decision-making for LLM agents.
Following prior work~\citep{kakade2002approximately,salimans2018learning,xiong2024watch}, we leverage expert trajectories as demonstrations and ask an LLM agent to begin exploration from the specific state $s_{t-1}\in\mathcal{S}$ toward the target state for a given demonstration. At each $t$-step, the agent's policy $\pi_\theta$ generates an action $a_t$, and we define a step-level reward $r_{step}(s_{t-1}, a_t)$ to quantify the contribution of $a_t$ to future success. 
Specifically, at $t$-th step, the agent generates $N$ new subsequent trajectories $\{e_{t+1:m}^{(i)}\}_{i=1}^{N}$ using the widely-used Monte Carlo sampling, conditioned on the historical trajectory $e_t$. Each trajectory receives an outcome reward $r_o(u,e_m)$ from the environment. The step-level reward $r_{step}(s_{t-1}, a_t)$ is computed as the expected value of these outcome rewards:
\begin{equation}
    r_{step}(s_{t-1}, a_t) = \mathbb E_{e_m \sim \pi_{\theta}(e_{t+1:m}|e_{t})} [r_o(u, e_m)].
\end{equation}

\paragraph{Normalized Dynamic Time Warping.}
\label{sec:distance_measure}
The normalized Dynamic Time Warping (nDTW) algorithm~\cite{muller2007dynamic}, implemented via dynamic programming (DP), effectively measures the distance between two trajectories containing multiple time steps. Formally, given a pair of trajectories $(x, y)$, this computation process is computed as:
\begin{align}
    D(i,j) = d(x_i, y_j) + 
        \min \begin{cases} 
            D(i-1, j) \\ 
            D(i, j-1),\\ 
            D(i-1, j-1)
        \end{cases}
\label{eq:nDTW}
\end{align}
where $d(x_i,y_i)$ denotes a distance function such as $L_2$ or cosine distance, $D(0,0)=d(x_0,y_0)$. $x_i$ denotes the action at the $i$-th step in the trajectory $x$, while $y_j$ denotes the action at the $j$-th step in the trajectory $y$. With a normalization operation, the nDTW distance $d_{\text{nDTW}}$ is given by:
\begin{equation}
    d_{\text{nDTW}}(x,y) = \frac{D(x-1, y-1)}{\sqrt{n_{x}^2 + n_{y}^2}},
\end{equation}
where $d_{\text{nDTW}}\in [0,1]$, $n_{x}$ and $n_{y}$ denote the number of steps in the trajectory $x$ and $y$, respectively. 

\section{Method}

In this section, we present \textbf{S}tep-level \textbf{T}raj\textbf{e}ctory \textbf{Ca}libration (\textbf{\model}), a novel learning framework for LLM agents.
First, we warm up agent training with supervised fine-tuning (\S\ref{sec3.1}), equipping LLM agents with necessary task planning capabilities. 
Then, we focus on calibration trajectory construction (\S\ref{sec3.2}), which detects deviated actions for an explored trajectory through step-level reward comparison and calibrates them by reflection. 
Finally, we utilize these calibrated trajectories as a crucial part of data for reinforced training (\S\ref{sec3.3}).
Figure~\ref{fig:overview} illustrates the overview of \model.

\subsection{Warm-up via Supervised Fine-tuning}
\label{sec3.1}

Supervised fine-tuning (SFT) on the expert trajectory data has demonstrated promising results, serving as an effective initial step for developing strong agents. We employ ReAct-style~\cite{yao2023react} trajectory to conduct SFT, which additionally generates a Chain-of-Thought (CoT)~\cite{wei2022chain} rationale before each action. Considering that the CoT and the corresponding action are generated together, we represent both as a single unit, denoted as $a_t$, for simplicity.
Given an expert trajectory dataset $\mathcal D = \Big\{(u, e)^{(i)}\Big\}_{i=1}^{|\mathcal D|}$, where each trajectory $e = (u, a_1, o_1, ..., a_m, o_m)$, $u$ represents the initial task instruction, $a_t$ denotes the action (including its rationale) at step $t$, $o_t$ is the corresponding observation, and $|\mathcal D|$ is the number of trajectories, the SFT loss function is formulated as:
\begin{equation}
    \mathcal L_{\text{SFT}}(\theta) = -\mathbb E_{e \sim \mathcal D}\bigg[\sum_{t=1}^n \log \pi_\theta(a_t|e_{t-1})\bigg].
\end{equation}
This warm-up process equips the LLM agent with the necessary task-planning capabilities, enabling it to generate both rationales and actions, resulting in a base agent $\pi_{\text{base}}$.

\subsection{Calibration Trajectory Construction}
\label{sec3.2}

To construct the calibration trajectories, we utilize the base agent $\pi_{\text{base}}$ to explore the environment through interaction. During this exploration, suboptimal actions often lead to a cascade of further suboptimal decisions, causing the trajectory to deviate from successful task completion. 
We define these actions, which are likely to cause deviations from the optimal trajectory and increase the risk of task failure, as \textit{deviated actions}. 
Below, we introduce the details of detecting deviated actions and constructing calibrated trajectories accordingly.

\paragraph{Deviated Action Detection via Step-level Reward Comparison.}

Since long-horizon tasks can be modeled as a partially observable Markov decision process (POMDP), where the future action in a task execution process depends on the current action, we must consider this Markov property when detecting deviated actions. 
To illustrate this, we define the probability of an agent successfully completing a task based on a ``good'' historical trajectory (e.g., an expert trajectory) at time step $t$ as $p(a_{>t}|a_{\leq t})$, where we omit the environmental states for simplicity.
After executing a subsequent action $a_{t+1}$, the probability of task completion becomes $p(a_{>t+1}|a_{\leq t+1})$. If $a_{t+1}$ is a ``good'' action (e.g., an expert action), $p(a_{>t+1}|a_{\leq t+1})$ will generally be greater than $p(a_{>t}|a_{\leq t})$. 
This is because agentic tasks typically consist of sequential actions, where each action contributes to task completion as the sequence progresses. 
Thus, by comparing the task completion probabilities before and after executing an action, we can determine whether the action is deviated.

Specifically, we employ step-level rewards, calculated via Monte Carlo (MC) sampling introduced in \S\ref{sec:task_form}, as an approximate estimation of the task completion probabilities.
The base agent would conduct an explored action $\hat{a}_{t+1}$ based on the expert sub-trajectory $e_{1:t}$.
The explored action $\hat{a}_{t+1}$ is classified as a deviated action if its step reward is significantly lower than that of the previous expert action $a_t$ by a predefined threshold $\delta$; otherwise, it is considered a non-deviated action.
The formal detection criterion is defined as follows:
\begin{equation}
    \begin{cases}
    \text{Deviated Action:} & \\
    \quad r_{step}(s_{t}, \hat{a}_{t+1}) - r_{step}(s_{t-1}, a_{t}) < \delta, \\
    \text{Non-deviated Action:} & \\
    \quad r_{step}(s_{t}, \hat{a}_{t+1}) - r_{step}(s_{t-1}, a_{t})\geq \delta,
    \end{cases}
\end{equation}
where $r_{step}(s_{t-1}, a_{t})$ represents the step reward for the expert action $a_{t}$ at the $t$-th step, $r_{step}(s_{t},\hat{a}_{t+1})$ denotes the step reward for the explored action $\hat{a}_{t+1}$, and $\delta \geq 0$ is a threshold parameter. Step rewards $r_{step}$ constructed by MC sampling are only utilized to detect the deviated actions.

\paragraph{Calibrated Trajectory Collection with Reflective Thoughts.}
As shown in Figure~\ref{fig:overview}, after identifying a deviated action in an explored trajectory, our goal is to enable the LLM agent to ``know'' that the action is deviated and learn how to realign with the task objective. Achieving this goal requires calibrated trajectories for training the agent. Inspired by many previous studies on LLM reflections~\citep{shinn2023reflexion}, we employ off-the-shelf  LLMs to generate reflective thoughts for calibration. Formally, we concatenate the preceding expert sub-trajectory $e_{1:t-1}$, the deviated action $\hat{a}_t$, and the corresponding ground-truth action $a_t$ in the expert trajectory, and prompt a state-of-the-art LLM (e.g., GPT-4o~\citep{openai2024gpt4o}) for reflection, transforming the deviated action $\hat{a}_t$ into the ground-truth action along with its reflective thought, which is denoted as $a_t^{'}$. This formulates the subsequent \textit{calibrated trajectory} $e_{c(t:m)} = (a_t^{'}, e_{t+1:m})$, where $e_{t+1:m}$ represents the expert sub-trajectory from the step $t+1$ to the end step $m$. The detailed prompt for this reflection is provided in Appendix~\ref{ref:reflection prompt}. 

Our calibration dataset $\mathcal{D}_c$ is constructed as:
\begin{equation}
    \mathcal{D}_c = \{e_{c(t:m)}^{(j)}\} \cup \{e_{d(1:m)}^{(j)}\},
\end{equation}
where $e_{d(1:m)}=(e_{1:t-1}, \hat{a}_t, \hat{e}_{t+1:m})$ denotes a deviated trajectory, which will be used in subsequent reinforced training.
Note that we perform trajectory calibration immediately when detecting the first deviated action, rather than waiting until the trajectory concludes. This approach ensures timely calibration and reduces unnecessary exploration.

\subsection{Reinforced Training}
\label{sec3.3}


While training on calibration trajectories enhances an agent's calibration capability, relying exclusively on these trajectories may hinder their ability to recognize correctness. To mitigate this, we introduce two types of successful data during exploration.
First, we construct the \textit{explored successful trajectory} dataset, $\mathcal{D}_e$, by collecting successful trajectories that the base agent independently explores from the beginning, along with their corresponding expert trajectories.
Not that these explored successful trajectories are not completely the same as the expert trajectories, because they include trial-and-error actions during explorations.
Second, we build the \textit{expert sub-trajectory} dataset, $\mathcal{D}_s$. Specifically, for a failed trajectory, where the first erroneous action occurs at step $t$, we extract the corresponding expert action and the subsequent trajectory, following \citet{xiong2024watch}. These sub-trajectories guide the agent in learning from challenging cases more effectively.

Using the collected data, we perform reinforced training to enhance LLM agents. 
Our goal is to guide the agent toward generating optimal trajectories that maximize task performance while minimizing suboptimal outcomes. We introduce \textit{trajectory deviation distance} (TDD), a measure that quantifies how much a suboptimal trajectory deviates from an optimal one at the trajectory level. 
Drawing inspiration from ~\citet{xu2024flame}, we utilize the nDTW distance $d_{\text{nDTW}}$ (as detailed in \S\ref{sec:distance_measure}), to quantify the deviation distance between a suboptimal trajectory $e_s$ and its corresponding optimal trajectory $e_o$.
A smaller $d_{\text{nDTW}}(e_s,e_o)$ indicates a lower deviation. 
This deviation distance will be utilized as a trajectory-level reward signal in reinforced training.

To ensure balanced training across the datasets, we refine the reward mechanism by incorporating the trajectory deviation distance.
The reward functions for each type of data are defined as follows:
\begin{align}
 r_c &= 1+\eta \cdot d_{\text{nDTW}}(e_{c(t:m)}, {e}_{d(t:m)}), \\
 r_{s} &= 1+\eta \cdot d_{\text{nDTW}}(e_{t:m}, \hat{e}_{t:m}),  \\
  r_{e} &= 1-\eta \cdot d_{\text{nDTW}}(\tilde{e}_{1:m}, e_{1:m}), 
\end{align}
where for the calibration trajectory $e_{c(t:m)}$ and the expert sub-trajectory $e_{t:m}$, we increase the reward as the deviation distance grows, encouraging the agent to calibrate larger deviations.
For the explored successful trajectory $\tilde{e}_{1:m}$, we reduce the reward for unnecessary explorations when the deviation distance increases, discouraging deviations from optimal behavior.
$\eta$ is a temperature coefficient that controls the impact of deviation distance on the reward.
Finally, we integrate these rewards into reinforcement training using the policy gradient~\citep{peters2007reinforcement} algorithm. The overall training objective is given by:
\begin{align}
& \mathcal{L}(\theta) = \nonumber \\
& \mathbb{E}_{(e_{c(t:m)}, e_{1:t-1}) \sim \mathcal{D}_c}
\Big[
    r_c \cdot \log \pi_\theta(e_{c(t:m)} \mid e_{1:t-1})
\Big] \nonumber \\
& + \mathbb{E}_{(e_{t:m}, e_{1:t-1}) \sim \mathcal{D}_{s}}
\Big[
    r_{s} \cdot \log \pi_\theta(e_{t:m} \mid e_{1:t-1})
\Big] \\
& + \mathbb{E}_{(\tilde{e}_{1:m}, u) \sim \mathcal{D}_{e}}
\Big[
    r_{e} \cdot \log \pi_\theta\big(\tilde{e}_{1:m} \big| u\big)
\Big] \nonumber.
\end{align}

\section{Experiments}

\subsection{Experimental Settings}

\paragraph{Datasets.}

We conduct experiments on two representative agentic task datasets: \textbf{VirtualHome}~\citep{puig2018virtualhome} and \textbf{ALFWorld}~\citep{shridhar2020alfworld}. 
For ALFWorld, we utilize datasets constructed by \citealp{song2024trial}. For the VirtualHome dataset, we leverage the predefined tasks from the ActivityPrograms knowledge base \citep{puig2018virtualhome} and construct a corresponding dataset in a manner closely aligned with the ALFWorld dataset.
Please refer to Appendix~\ref{ref:ds} for further details regarding the dataset construction process and associated statistical information.

\begin{table*}[t!]
    \centering
    \resizebox{0.98\textwidth}{!}{
    \begin{tabular}{l l c c c c c}
    \toprule
    \multirow{2}{*}{\textbf{Paradigm}} & \multirow{2}{*}{\textbf{Method}} & \multicolumn{2}{c}{\textbf{VirtualHome}} & \multicolumn{2}{c}{\textbf{ALFWorld}} & \multirow{2}{*}{\textbf{Average}}\\
    \cmidrule(l){3-4} \cmidrule(l){5-6}
         & & Seen & Unseen & Seen & Unseen & \\
    \midrule
    \multirow{3}{*}{Prompting-based} 
    & GPT-3.5-Turbo~\citep{ouyang2022training} & 6.3 & 2.6 & 7.9 & 10.5 & 6.8 \\
    & GPT-4~\citep{achiam2023gpt} & 34.2 & 9.4 & 42.9 & 38.1 & 31.2 \\
    \midrule
    \multirow{10}{*}{Tuning-based} 
    & Llama-2-7B-Chat + PPO~\citep{schulman2017proximal} & 23.9 & 25.0 & 22.1 & 29.1 & 25.0 \\
    & Llama-2-7B-Chat + SFT~\citep{chen2023fireact} & 64.9 & 57.7 & 60.0 & 67.2 & 63.3 \\
    & Llama-2-7B-Chat + RFT~\citep{yuan2023scaling} & 65.1 & 58.3 & 62.9 & 66.4 & 63.2 \\
    & Llama-2-7B-Chat + Step-PPO~\citep{wang-etal-2024-math} & 65.7 & 59.6 & 65.7 & 69.4 & 65.1 \\
    & Llama-2-7B-Chat + ETO~\citep{song2024trial} & 66.6 & 60.1 & 68.6 & 72.4 & 66.9 \\
    & Llama-2-7B-Chat + E$^2$CL~\citep{wang-etal-2024-e2cl} & 67.1 & 61.8 & 70.1 & 73.9 & 68.2 \\
    & Llama-2-7B-Chat + IPR~\citep{xiong2024watch} & 67.6 & 61.9 & 70.3 & 74.7 & 68.6 \\
    \cmidrule{2-7}
    & \cellcolor[gray]{0.9}\textbf{Llama-2-7B-Chat + STeCa (Ours)} & \cellcolor[gray]{0.9}\textbf{69.6} & \cellcolor[gray]{0.9}\textbf{63.6} & \cellcolor[gray]{0.9}\textbf{74.3} & \cellcolor[gray]{0.9}\textbf{76.1} & \cellcolor[gray]{0.9} \textbf{70.9} \\
    & \textbf{Llama-2-7B-Chat + STeCa w/ SFT+DPO} & 66.8 & 63.5 & 74.1 & 75.5 & 70.0 \\
    & \textbf{Llama-2-7B-Chat + STeCa w/o RT} & 68.8 & 62.4 & 72.1 & 74.9 & 69.6 \\
    \bottomrule
    \end{tabular}
    }
    \caption{Performance of different methods on two agent datasets. ``Seen'' refers to the held-out test set containing tasks present during training, while ``Unseen'' refers to the test set with unseen task variations.}
    \label{tab:main_results}
\end{table*}

\paragraph{Baseline Methods.}

We evaluate STeCa against the following two categories of baselines: (1) \textbf{prompting-based} approaches, including GPT-3.5-turbo~\citep{ouyang2022training} and GPT-4~\citep{achiam2023gpt}. 
(2) \textbf{tuning-based} methods, which include supervised fine-tuning (SFT) methods, such as pure SFT~\citep{chen2023fireact}, RFT~\citep{yuan2023scaling}, and E$^2$CL~\citep{wang-etal-2024-e2cl}, reinforcement learning-based methods such as PPO~\citep{schulman2017proximal} and Step-PPO~\citep{wang-etal-2024-math}, as well as preference learning methods like ETO~\citep{song2024trial} and IPR~\citep{xiong2024watch}. Additional details about the baselines are provided in Appendix~\ref{appendix:baselines}.

\paragraph{Implementation Details.}

We utilize Llama-2-7B-Chat~\citep{touvron2023llama} as the base model for training LLM agents. We set $\delta=0$ as the threshold to detect deviated actions in two environments. We use $\eta=1$ for VirtualHome and $\eta=0.01$ for ALFWorld to weight the contribution of trajectory deviation to the reward. To obtain step-level rewards with MC sampling, we set the temperature to 1 and the number of samples $N$ to 5. More details are presented in Appendix~\ref{appendix:implement}.

\paragraph{Evaluation Metrics.}
Following existing studies~\cite{song2024trial,xiong2024watch}, we adopt the \textbf{Average Final Reward} as our evaluation metric. This metric measures the success rate of test tasks. 
In ALFWorld, the environment provides binary final rewards, where a reward of 1 indicates task completion and 0 indicates failure. 
Similarly, in VirtualHome, a trajectory is deemed successful if the final environment state aligns with a predefined target state and yields a reward of 1; otherwise, the reward is 0. 
This metric ensures a consistent measure of task performance across both environments.

\subsection{Main Results}
Table~\ref{tab:main_results} summarizes the performance of various methods on long-horizon tasks in the VirtualHome and ALFWorld environments. The proposed method, STeCa, achieves the highest overall performance, with an average reward of 70.9, significantly outperforming the baseline methods.
Compared to prompting-based methods, which exhibit relatively poor performance, STeCa demonstrates a significant improvement.
These results highlight the inherent limitations of closed-source LLMs relying solely on prompt engineering.
As a tuning-based method, STeCa demonstrates consistent superiority over prior approaches. Specifically, it achieves an average final reward of 70.9, surpassing IPR, the previous state-of-the-art method with an average reward of 68.6, by 3.4\%. This improvement highlights the effectiveness of trajectory calibration in enhancing generalization and overall performance. Moreover, STeCa outperforms E$^2$CL, a method that incorporates self-reflection mechanisms, by 4.0\%. 
Notably, STeCa achieves this without requiring additional iterative training, underscoring its superior training efficiency.

To further validate our method, we conducted an ablation study by designing two variants of the training process. In the first variant, we applied supervised fine-tuning (SFT) followed by Direct Preference Optimization (DPO)~\citep{rafailov2024direct} on both optimal and suboptimal trajectories in the collected dataset (denoted as w/ SFT + DPO). In the second variant, we performed SFT on the collected dataset but restricted the training to only the optimal trajectories, omitting reward tuning (denoted as w/o RT).
The experimental results showed that employing both SFT and DPO led to a slight decrease in the average reward to 70.0, while SFT without reward tuning (w/o RT) resulted in a further drop to 69.6. Although both variants demonstrated competitive performance, neither surpassed the results achieved by STeCa, thereby underscoring the effectiveness of the mechanisms proposed in our method.

\begin{table}[t!]
    \centering
    \resizebox{0.85\linewidth}{!}{
    \begin{tabular}{l c c c}
    \toprule
    \textbf{Base Model}  &  \textbf{Method} & \textbf{Seen} & \textbf{Unseen}\\ \midrule
    \multirow{3}{*}{Mistral-7B}  & SFT & 67.1 & 58.2 \\
    & IPR & 71.4 & 73.9 \\
    \cmidrule{2-4}
    & \textbf{STeCa} & \textbf{73.3} & \textbf{75.3}  \\ 
    \midrule
    \multirow{3}{*}{Llama-3-8B-Instruct} & SFT & 68.6 & 62.7 \\
    & IPR & 72.3 & 75.8 \\
    \cmidrule{2-4}
    & \textbf{STeCa} & \textbf{74.9} &  \textbf{77.0}  \\
    \bottomrule
    \end{tabular}
    }
    \caption{Performance of SFT, IPR, and our \model with different base models on the ALFWorld dataset.}
    \label{tab:base model}
\end{table}

\section{Discussions and Analyses}
In this section, we provide detailed analyses of the \model framework from the following aspects.

\subsection{Effectiveness with Different Base Models}

To validate the broad effectiveness of our method, we evaluate STeCa with different base models, including Mistral-7B and Llama-3-8B-Instruct, on the ALFWorld environment. We compare its performance with SFT and IPR across both seen and unseen tasks. As shown in Table~\ref{tab:base model}, STeCa consistently outperforms both SFT and IPR across multiple base models. 
Notably, in unseen tasks, STeCa achieves a 17.1\% improvement over SFT on Mistral-7B, highlighting its generalization ability for developing LLM-based agents. 
Furthermore, with Llama-3-8B-Instruct, a stronger backbone model, STeCa further achieves better performance, underscoring its potential for building advanced agents in the future.

\subsection{Comparisons between Variants of STeCa}
\label{analyses: ablation study}

\paragraph{Variants of Step-Level Reward Acquisition.}

To investigate the impact of different methods for reward acquisition, we conducted experiments using GPT-4o and a trained reward model to annotate reward for each step action while keeping all other processes unchanged. 
The detailed process of step action reward acquisition is described in Appendix~\ref{appendix:reward_analysis}. 
As shown in Table~\ref{tab:ablation}, our method utilizing MC sampling for step reward acquisition achieves superior performance compared to alternative variants, highlighting the effectiveness of MC sampling for reward acquisition. Notably, employing GPT-4o to directly annotate rewards for step actions demonstrates performance comparable to our method, suggesting that step rewards can be effectively obtained through more computationally efficient approaches. This finding provides a promising direction for future research into optimizing the efficiency of reward acquisition.

\paragraph{Variants of Reflective Thought Generation.}

In STeCa, we employ GPT-4o to generate reflection thoughts for constructing calibration trajectories.
To evaluate the impact of reflection quality on performance, we attempted to prompt the base agent $\pi_{base}$ to generate reflections while keeping all other processes unchanged. The results, summarized in Table~\ref{tab:ablation}, reveal a significant performance degradation, underscoring the critical importance of high-quality reflection generation for achieving optimal results. This finding also suggests that the base agent, trained solely on expert trajectories, lacks effective reflection capabilities.

\begin{table}[t!]
\centering
\resizebox{0.95\linewidth}{!}{
\begin{tabular}{lccccc}
\toprule
\multirow{2}{*}{\textbf{Variants}}&  \multicolumn{2}{c}{\textbf{VirtualHome}} & \multicolumn{2}{c}{\textbf{ALFWorld}} \\
\cmidrule(l){2-3}
\cmidrule(l){4-5}
      & Seen & Unseen & Seen & Unseen \\
\midrule
\multicolumn{5}{c}{\textbf{Step-level Reward Acquisition}} \\
\midrule
MC Sampling  &  69.6 &  63.6  & 74.3  & 76.1   \\
GPT-4o Annotation  &  69.1  &  62.5 &  74.1 & 74.9  \\
RM Prediction   &  68.2 &  61.8 &  74.0  & 73.3   \\
\midrule
\multicolumn{5}{c}{\textbf{Reflective Thought Generation}} \\
\midrule
GPT-4o Generation      &  69.6  &  63.6 &  74.3   & 76.1   \\
Self-generation      &  66.0  &  61.1 &  71.4  & 73.3   \\
\bottomrule
\end{tabular}
}
\caption{Comparisons between variants of \model.}
\label{tab:ablation}
\end{table}

\begin{figure}[t!]
  \includegraphics[width=1.0\linewidth]{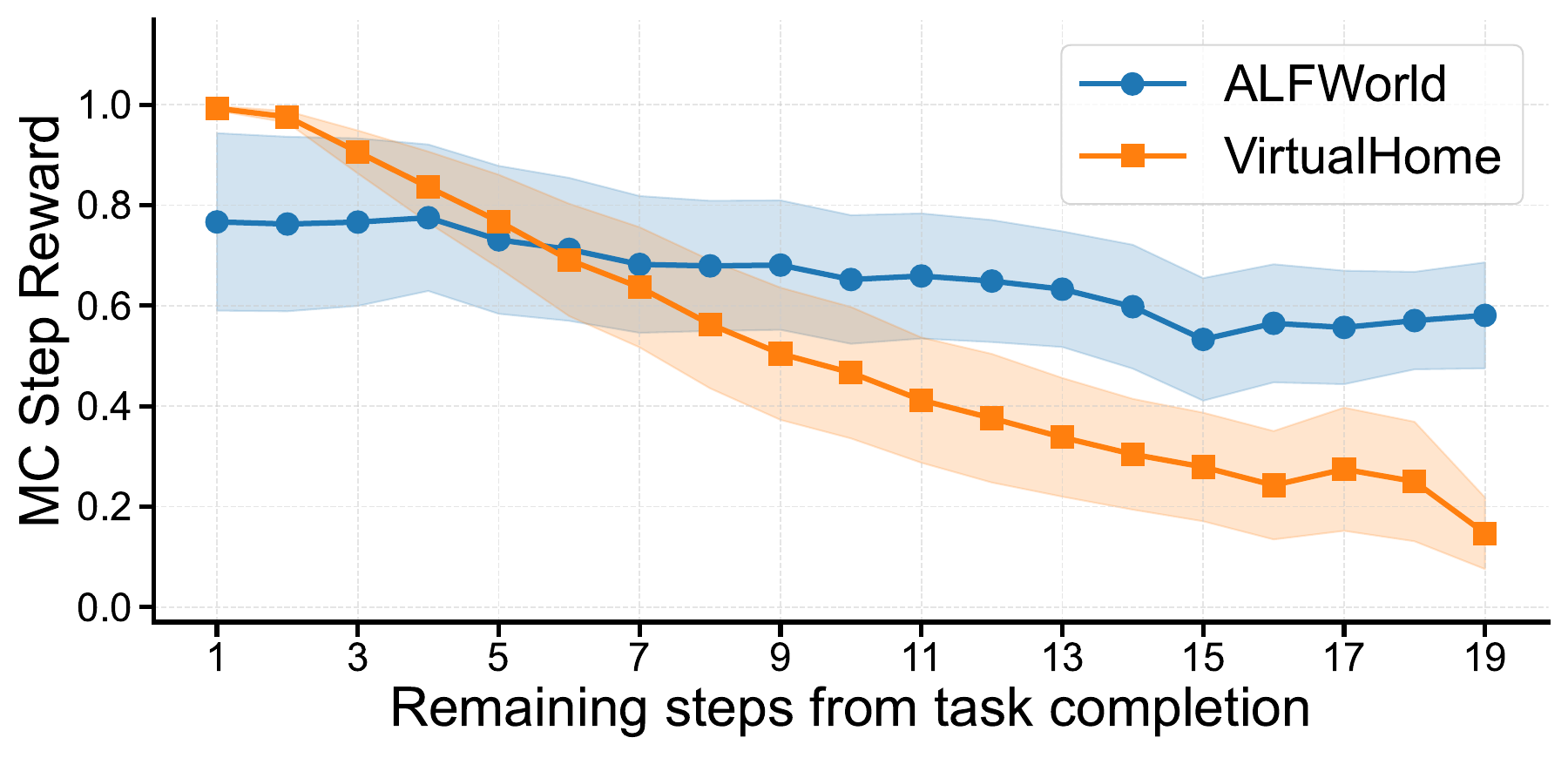}
  \caption{Variations in Monte Carlo (MC) step rewards with respect to the number of remaining steps until task completion for expert trajectories.}
  \label{fig:average_rewards_comparison}
\end{figure}

\subsection{Analyses of Deviated Action Detection}
\label{statistic of step rewards}

To validate the empirical Markov property introduced in Section~\ref{sec3.2}, i.e., non-deviated ``good'' actions increase the likelihood of task completion, we conducted a statistical analysis using expert trajectories. Specifically, we compared task completion probabilities at varying distances from task completion, employing MC step rewards as a proxy for these probabilities. 
As illustrated in Figure~\ref{fig:average_rewards_comparison}, MC step rewards monotonically increase as the agent progresses toward task completion in both environments. This trend demonstrates that the accumulation of optimal actions significantly contributes to task completion. 
Conversely, deviated actions consistently reduce the task completion probability, further supporting our approach of using step reward comparisons between adjacent steps as a reliable criterion for detecting deviated actions.

\subsection{Analyses of Calibration}

\begin{figure}[t!]
  \includegraphics[width=1.0\linewidth]{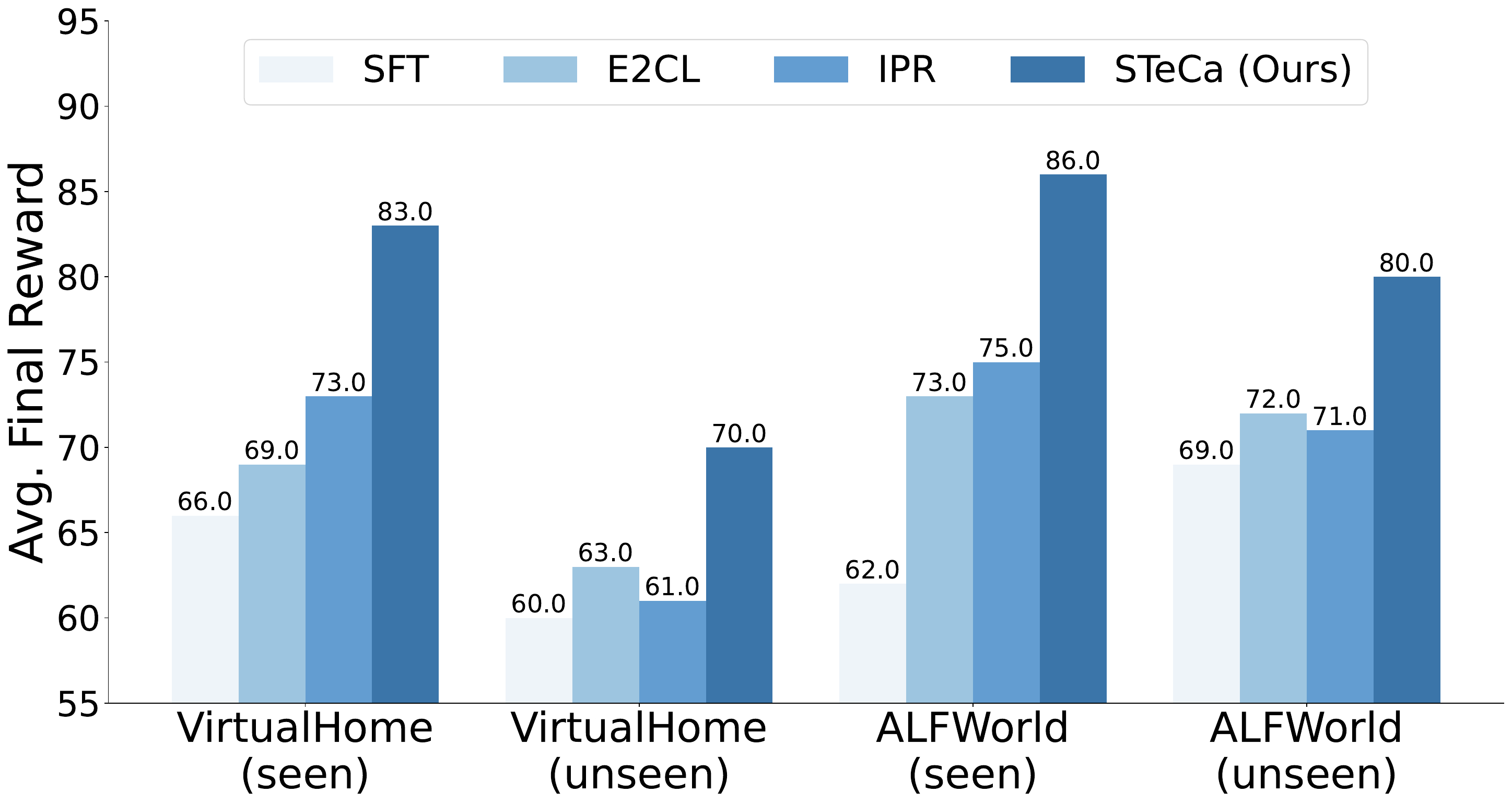}
  \caption{Calibration performance of different methods on the VirtualHome and ALFWorld datasets.
  }
  \label{fig:calibration_performance}
\end{figure}

In this section, we compare the calibration capabilities of STeCa and baseline methods. We evaluate calibration performance using the average final reward achieved upon successful task completion in the presence of deviated actions. To enable this analysis, we constructed datasets containing historical trajectories with deviated actions, categorized into seen and unseen scenarios across both environments. Additional details are provided in Appendix~\ref{appendix:calibration_analysis}.
As shown in Figure~\ref{fig:calibration_performance}, STeCa outperforms baseline methods by a significant margin. For instance, it achieves a 14.8\% relative improvement over IPR on unseen tasks in the VirtualHome environment, highlighting its superior calibration performance. To comprehensively illustrate the performance of different agent learning methods, we provide some concrete examples in Appendix~\ref{appendix:case_study}.

\begin{figure}[t!]
  \includegraphics[width=1.0\linewidth]{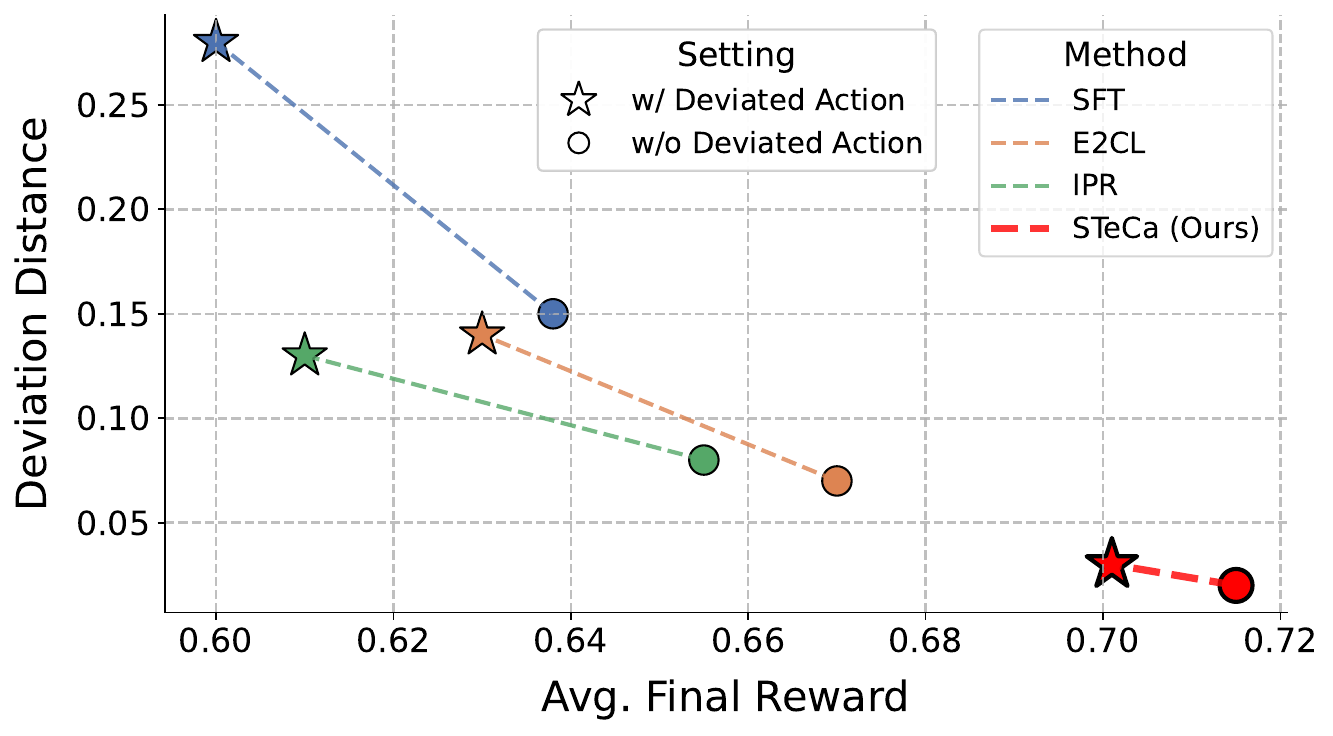}
  \caption{Correlation between the deviation distance and success rate (measured by average final reward).
  }
  \label{fig:deviation_success_rate}
\end{figure}

To examine the impact of deviated actions on LLM agent performance, we evaluate these agents trained with different methods under two settings: one with deviated actions included in historical trajectories and another without. The construction process of the testing datasets is detailed in Appendix~\ref{appendix:calibration_analysis}.
As shown in Figure~\ref{fig:deviation_success_rate}, STeCa exhibits minimal performance variation between the two settings, unlike other methods, which show significant differences. 
This suggests that the presence of deviated actions has little effect on STeCa's performance, as it effectively calibrates subsequent trajectories. 
These findings highlight STeCa's robustness and its potential for reliable performance in diverse and challenging environments.

\section{Related Work}

\paragraph{LLM Agent Learning.}
LLM agents are widely used for tackling complex real-world tasks~\citep{wang2023voyager,hu2024dawn,wang-etal-2024-e2cl}, relying on iterative interactions with their environment guided by task objectives and constraints. However, in long-horizon planning, excessive interactions make them prone to suboptimal actions, increasing the risk of failure. While closed-source LLMs demonstrate strong intelligence, open-source counterparts still lag behind~\citep{liu2023agentbench,wang2023mint}.
To address this gap, some studies focus on improving task success rates by increasing the likelihood of generating optimal actions~\citep{chen2023fireact, yuan2023scaling}. Alternatively, another line of research seeks to mitigate suboptimal actions by collecting them and applying preference learning methods to reduce their occurrence~\citep{song2024trial,xiong2024watch}.
Recently, researchers have explored the capacity of LLM agents to self-correct errors, enhancing their ability to ensure successful task completion~\citep{wang-etal-2024-e2cl, qu2024recursive}.
However, these methods primarily focus on self-correction after errors have already occurred, lacking the ability to detect suboptimal actions in advance and calibrate subsequent planning accordingly.

\paragraph{Process Supervision.}
 
Process supervision provides fine-grained guidance, making it a promising approach for addressing long-horizon problems~\citep{uesato2022solving}. Early studies have explored obtaining step-level rewards and using them to optimize intermediate processes through reinforcement learning~\citep{lightman2023let, deng2024novice, wang-etal-2024-math}. Others have focused on constructing step-level positive and negative data pairs and applying preference learning techniques to achieve more precise optimization~\citep{xiong2024watch, jiao2024learning}.
However, existing studies have yet to address the construction of step-level reflection data. Such data could empower LLM agents to detect suboptimal actions, analyze the reasons for their suboptimality, and determine how to calibrate them to ensure successful task completion.

\section{Conclusion}

In this paper, we introduce STeCa, a novel agent learning framework designed to enhance the performance of LLM agents in long-horizon tasks. 
STeCa identifies deviated actions through step-level reward comparisons and constructs calibration trajectories via reflection. 
These trajectories serve as critical data for reinforced training. Extensive experiments demonstrate that STeCa significantly outperforms baseline methods, with additional analyses underscoring its robust calibration capabilities.


\section*{Limitations}

While our approach demonstrates superior performance compared to baseline methods, it is important to acknowledge the limitations of our current work as follows:

(1) Computational Inefficiency: Although the Monte Carlo (MC) sampling approach in STeCa achieves superior performance in constructing step rewards compared to alternative methods, it requires a substantial number of sampling iterations, resulting in significant computational overhead. This inefficiency represents a notable limitation of our current implementation. Future work should focus on developing more efficient methods for constructing step rewards while preserving the performance advantages of our approach.

(2) Limited Utilization of Step Rewards: While our approach leverages step rewards to identify and evaluate deviated actions effectively, it does not fully exploit the potential of step rewards for broader decision-making or optimization tasks. This constrained utilization may limit the overall performance improvements that could be achieved by incorporating step rewards into other aspects of the framework. Future research should explore strategies to better harness the rich information embedded in step rewards to enhance the overall effectiveness and adaptability of the system.

(3) Handling Multiple Deviated Actions: Our method primarily focuses on ``timely'' calibration, wherein the LLM agent identifies a deviated action and immediately adjusts its subsequent trajectory to prevent further deviations. This approach effectively mitigates the accumulation of errors over time. Nonetheless, our current framework does not explicitly address multi-step calibration for multiple deviated actions. Systematically handling such cases presents an opportunity for future work, which could further enhance the robustness of our method in more complex scenarios.

\section*{Ethics Statement}

This work aims to develop LLM agents within simulated environments. The VirtualHome and ALFWorld environment setup and related data strictly follow the specifications of VirtualHome \citep{puig2018virtualhome} and ALFWorld \citep{shridhar2020alfworld}. We utilize VirtualHome v2.3.0\footnote{\url{https://github.com/xavierpuigf/virtualhome/tree/master}} (MIT license\footnote{\url{https://github.com/xavierpuigf/virtualhome/blob/master/LICENSE}}) and ALFWorld\footnote{\url{https://github.com/alfworld/alfworld}} (MIT license\footnote{\url{https://github.com/alfworld/alfworld/blob/master/LICENSE}}) to conduct our experiments. 
All the LLMs we use for fine-tuning are open-source, and we strictly follow the protocols for the academic use of these models.
Additionally, we acknowledge the use of AI assistants, including GitHub Copilot and ChatGPT, in supporting our coding and writing processes.

\section*{Acknowledgements} 
This work was supported by the Research Grants Council of Hong Kong (15209724). The authors would like to thank the anonymous reviewers for their valuable feedback and constructive suggestions.


\bibliography{custom}

\appendix
\newpage

\clearpage
\newpage
\section{Datasets and Preprocessing}
\label{ref:ds}

\paragraph{ALFWorld}
ALFWorld~\citep{shridhar2020alfworld} offers interactive TextWorld environments that are meticulously aligned with the embodied environments introduced in ALFRED~\citep{shridhar2020alfred}. This framework challenges agents to navigate complex household settings and execute high-level instructions, thereby testing their ability to perform practical tasks. The dataset is structured into two distinct evaluation sets: a seen set, designed to assess in-distribution generalization, and an unseen set, which comprises novel task instances to evaluate out-of-distribution generalization capabilities. At the conclusion of each trajectory, the environment provides a binary reward, indicating whether the agent has successfully completed the assigned task. This setup facilitates a clear and measurable assessment of agent performance in both familiar and novel scenarios.

\paragraph{VirtualHome} 

VirtualHome~\citep{puig2018virtualhome} is a comprehensive dataset comprising 292 high-level household tasks and 1,374 unique action plans, distributed across 6,201 diverse environments. The dataset was meticulously curated through manual annotations provided by Amazon Mechanical Turk workers, who labeled tasks and their corresponding action plans in detail. Each entry in the dataset is structured into three components: a high-level task, a descriptive explanation, and executable action programs compatible with the VirtualHome environment. To evaluate task completion, we executed all tasks and recorded the final state of the environment upon completion. A task is considered successfully completed if the state of the environment after exploration by the LLM agent matches the predefined target state. To ensure data quality, the dataset was filtered by retaining only trajectories with successful final outcome rewards and verifying that every action in the planning sequence is executable within the environment. Furthermore, to maintain an appropriate level of task complexity, the dataset was restricted to trajectories with planning lengths ranging from 3 to 20 steps. This rigorous filtering process ensures a robust and reliable subset of data, suitable for in-depth analysis and model training.

\paragraph{Dataset Construction}
Since the original trajectories do not include reasoning processes preceding each action, we adopt established methodologies from prior work~\citep{song2024trial,xiong2024watch} to enrich the data. Specifically, we incorporate relevant task information and expert action trajectories to prompt GPT-4o to generate plausible reasoning steps (thoughts) before each action. This approach ensures that the dataset captures the cognitive processes underlying decision-making. Ultimately, the datasets are structured in a thought-action format, following the ReAct framework~\citep{yao2023react}. Detailed statistics for the two datasets are provided in Table~\ref{tab:dataset}, highlighting their key characteristics and composition.

\section{Baseline Methods}
\label{appendix:baselines}

Our baseline methods are as follows: 
1) SFT~\citep{chen2023fireact}, which employs behavior cloning on expert trajectories alone, serving as the base agent for STeCa and other baseline methods.
2) PPO~\citep{schulman2017proximal}, a widely-used reinforcement learning algorithm, optimizes final trajectory rewards. Additionally, we apply PPO for stepwise action optimization.
3) RFT~\citep{yuan2023scaling}, which extends expert trajectories by incorporating successful trajectories discovered by the base agent, followed by fine-tuning on the expanded dataset.
4) ETO~\citep{song2024trial}, which constructs positive and negative trajectory pairs and optimizes them using Direct Preference Optimization (DPO)~\citep{rafailov2024direct}.
5) E$^2$CL~\citep{wang-etal-2024-e2cl}, which leverages planning data, feedback data, and correction data to supervise the fine-tuning of LLM agents.
6) IPR~\citep{xiong2024watch}, which enhances trajectory pairs by augmenting sub-trajectory pairs based on step rewards, building upon ETO's framework, and trains LLM agents using preference learning methods.

\begin{table}[t!]
    \centering
    \resizebox{1.0\linewidth}{!}{
    \begin{tabular}{l c c c c c}
    \toprule
    \textbf{Dataset}   & \textbf{Train} & \textbf{Test} & \textbf{\#Actions}  & \textbf{\#Avg./Max. Turns}\\
    \midrule
    ALFWorld & 2,851 & 274 & 13 & 10.1 / 20 \\
    VirtualHome & 4,920 & 494 & 40 & 11.5 / 20\\
    \bottomrule
    \end{tabular}
    }
    \caption{Statistics of the datasets for experiments.}
    \label{tab:dataset}
\end{table}

\section{Additional Implementation Details}
\label{appendix:implement}

During the construction of the base agent, we train the model for 3 epochs with a batch size of 16 and a learning rate of 3e-6, employing the AdamW optimizer and a cosine learning rate scheduler. For reinforced training, the model is fine-tuned for only 1 epoch.

During the inference phase, all methods are evaluated using the ReAct-style interaction format, where the agent generates a rationale before executing each action. Specifically, we include a one-shot example in the instruction prompt for each task. Detailed prompts are provided in Appendix~\ref{ref:prompt}. For text generation, we apply greedy decoding with the temperature set to 0. To accelerate inference, we utilize vLLM~\citep{kwon2023efficient} libribray to optimize the generation process of LLMs. 

All experiments were conducted on a computational cluster equipped with 8 NVIDIA A6000 48GB GPUs. For fine-tuning, we employed several open-source models, including Llama-2-7B-Chat~\citep{touvron2023llama}, Mistral-7B~\citep{jiang2023mistral}, and Llama-3-8B-Instruct. We strictly complied with the licensing terms for academic use associated with these models: Llama-2-7B-Chat is governed by the Llama 2 Community License\footnote{\url{https://huggingface.co/meta-llama/Llama-2-7b-chat-hf/blob/main/LICENSE.txt}}, Mistral-7B is licensed under the Apache-2.0 License\footnote{\url{https://huggingface.co/mistralai/Mistral-7B-v0.1}}, and Llama-3-8B-Instruct adheres to the Llama 3 License\footnote{\url{https://huggingface.co/meta-llama/Meta-Llama-3-8B/blob/main/LICENSE}}. This adherence ensures that our use of these models aligns with their respective legal and ethical guidelines.

\section{Experimental Settings about Analyses}
\label{appendix:analysis}

\subsection{Variants of Step-level Reward Acquisition}
\label{appendix:reward_analysis}

In addition to the Monte Carlo (MC) sampling for step-level reward acquisition, we further employ the following two variants: 
(1) \textbf{GPT-4o Annotation:} In Section~\ref{sec3.2}, we collect Monte Carlo (MC) step rewards corresponding to various step actions. To annotate all explored step actions, we randomly select several samples as in-context examples and utilize GPT-4 for annotation. The detailed prompt used for this process is provided in Appendix~\ref{appdendix: reflection prompt for step rewards}.
(2) \textbf{Reward Model Prediction:} We also leverage the data collected in Section~\ref{sec3.2}, where each step action is associated with an MC step reward, to train a reward model capable of predicting scores for step actions. Specifically, we use the Llama-2-7B-Chat~\citep{touvron2023llama} model as the base architecture. To mitigate overfitting, we add a dropout layer to the output layer, followed by a linear layer to map the output to a scalar score. Additionally, we employ Low-Rank Adaptation (LoRA)~\citep{hu2021lora} for efficient fine-tuning. The model is trained for 3 epochs, and during testing, we set the random seed to 42 to ensure reproducibility and score all step actions.

\subsection{Detailed Settings for Calibration Analysis}
\label{appendix:calibration_analysis}

We randomly select 100 pieces of data from $D_c(e_{1:t-1}, \hat{a}_t, e_{c(t:m)}, \hat{e}_{t+1:m})$ for both ALFWorld and VirtualHome to serve as the seen test set. Additionally, we randomly select 100 pieces of data from the unseen test set. Following the procedure outlined in Section~\ref{sec3.2}, we construct the calibration dataset $(e_{1:t-1}, \hat{a}_t, e_{c(t:m)}, \hat{e}_{t+1:m})$ derived from unseen scenarios.
After assembling the calibration datasets for both seen and unseen scenarios in VirtualHome and ALFWorld, we use these datasets to evaluate the calibration performance of the LLM agent. Specifically, we traverse the step actions from $(e_{1:t-1}, \hat{a}_t)$ to obtain the initial environment state. We then deploy the LLM agent to explore the environment starting from this state and assess whether it can successfully complete the task.

For the second experiment, we reuse the previously collected calibration dataset. However, in this case, we traverse the step actions only from $e_{1:t-1}$, excluding the deviated action $\hat{a}_t$. We refer to this configuration as the ``w/o deviated action'' setting.

\section{Prompt Templates}
\label{ref:prompt}

\subsection{Inference Prompt}
\label{ref:inference prompt}

As shown in Figure~\ref{fig:Inference prompt}, we provide the inference prompt for each task, which include a general instruction, a one-shot example, the specific task instruction and history trajectory.

\begin{tcolorbox}[breakable,title=Inference Prompt]
\textbf{\textit{\# General Instruction:}} \\
\textbf{Human}: Interact with a household to solve a task. Imagine you are an intelligent agent in a household environment and your target is to perform actions to complete the task goal...\\
Your response should use the following format: \\
Thought: <your thoughts> \\
Action: <your next action> \\
\textbf{Agent}: OK\\

\textbf{\textit{\# In-Context Example:}} \\
\textbf{Human}: The task is Drink (Drink water). \\
... \\

\textbf{\textit{\# Task Instruction:}} \\
\textbf{Human}: The task is xxx. \\
\textit{\textbf{(History trajectory)}} \\
... \\

\end{tcolorbox}
\begin{figure}[ht]
    \centering
    \vspace{-8pt}
    \caption{Inference prompt template.}
    \label{fig:Inference prompt}
\end{figure}

\subsection{Reflection Prompt}
\label{ref:reflection prompt}

As shown in Figure~\ref{fig:reflection prompt for virtualhome}, the reflection prompt includes history trajectory (containing deviated action) and reflection instruction. This prompt is then used to request GPT-4o to generate reflective thoughts.

\begin{tcolorbox}[breakable,title=Reflection Prompt]
\textbf{\textit{\# Historical Trajectory:}} \\
\textbf{Human}: Interact with a household to solve a task. Imagine you are an intelligent agent in a household environment and your target is to perform actions to complete the task goal...\\
\textbf{Agent}: OK\\
\textbf{Human}: Your task is write an email... \\
\textbf{Agent}: Thought: ... Action: ... \\
... (Interaction with multi-turns)\\
\textbf{Agent}: Thought: ... Action: ... \textcolor{red}{(\# error action at this step)} \\

\textbf{\textit{\# Reflection Instruction:}} \\
Above is the interaction history. However, the last step is not optimal and may lead to a wrong direction. The next step ground-truth action is [\textcolor{blue}{\text{ground truth action at this step}}]. Please provide the thought which would lead the agent to generate the ground truth action and be aware of the last non-optimal action. The thought should follow the format of the interaction history.

\end{tcolorbox}
\begin{figure}[ht]
    \centering
    \vspace{-8pt}
    \caption{
    Prompt template for reflection.
    }
    \label{fig:reflection prompt for virtualhome}
\end{figure}

\subsection{Prompt for Step Reward Prediction}
\label{appdendix: reflection prompt for step rewards}
Figure~\ref{fig:reflection prompt for step rewards in vh} presents the prompt template designed for predicting step rewards, which consists of an instruction and several in-context examples.

\begin{tcolorbox}[breakable,title=Prompt for Step Reward Prediction]
\textbf{\textit{\# Instruction:}} \\
Based on the historical interaction trajectory, the agent has generated its final action. Evaluate whether this action is conducive to the completion of the ultimate task and provide a score. The score should range from 0 to 1, with scores closer to 1 indicating a higher likelihood of task completion. Generate only the score, without any additional commentary or explanation. \\

\textbf{\textit{\# History trajectory1:}} \\
xxx \\
\textbf{\textit{\# Score1:}} \\
xx \\
\textbf{\textit{\# History trajectory2:}} \\
xxx \\
\textbf{\textit{\# Score2:}} \\
xx \\
\textbf{\textit{\# History trajectory3:}} \\
xx \\
\textbf{\textit{\# Score3:}} \\
...
\end{tcolorbox}

\begin{figure}[ht]
    \centering
    \vspace{-8pt}
    \caption{
    Prompt template for step reward prediction.
    }
    \label{fig:reflection prompt for step rewards in vh}
\end{figure}

\section{Case study}
\label{appendix:case_study}

Figure~\ref{fig:alfworld_case} provides an example demonstrating STeCa's calibration capability during the planning process. As illustrated, STeCa autonomously identifies deviated actions, reflects on them, and successfully completes the task. In contrast, the other two agents fail to achieve the same level of performance, further highlighting the effectiveness of STeCa in addressing such challenges.

\onecolumn

\begin{tcolorbox}[breakable,title=Case Study on the ALFWorld Benchmark]
\columnseprule=0.5pt
\begin{multicols}{2}

\textcolor{case_purple}{\textbf{Task Instruction:}} Your task is to find two pans and put them on the countertop.\\
\textcolor{case_blue}{\textbf{Thought 1:}} I need to find two pans for the task. The first step is to search the cabinets as they are likely places where pans are stored.\\
\textcolor{case_blue}{\textbf{Action 1:}} go to cabinet 1\\
\textcolor{case_purple}{\textbf{Observation 1:}} On cabinet 1, you see: a plate 1, a bowl 1, and a glass 1.\\
\textcolor{case_blue}{\textbf{Thought 2:}} Cabinet 1 contains a plate, a bowl, and a glass, but no pan. I should check the next cabinet for a pan.\\
\textcolor{case_blue}{\textbf{Action 2:}} go to cabinet 2\\
\textcolor{case_purple}{\textbf{Observation 2:}} On cabinet 2, you see: a cuttingboard 1, a knife 1, and a pan 1.\\
\textcolor{case_blue}{\textbf{Thought 3:}} Cabinet 2 also doesn't have a pan, just a saltshaker. I'll continue searching the cabinets.\\
\textcolor{case_blue}{\textbf{Action 3:}} go to cabinet 3\quad \textcolor{red}{\textit{\textbf{(\# Deviated Action)}}}\\
\textcolor{case_purple}{\textbf{Observation 3:}} The cabinet 3 is closed.\\

\begin{center}\textbf{Agent w/ SFT}\end{center}
\textcolor{case_blue}{\textbf{Thought 4:}} Cabinet 3 is closed. I need to open it to see if there's a pan inside.\\
\textcolor{case_blue}{\textbf{Action 4:}} open cabinet 3\\
\textcolor{case_purple}{\textbf{Observation 4:}} You open the cabinet 3. The cabinet 3 is open. In it, you see a plate 2.\\
\textcolor{case_blue}{\textbf{Thought 5:}} Inside cabinet 3, there's only a plate. I should close this cabinet and keep looking for a pan.\\
\textcolor{case_blue}{\textbf{Action 5:}} close cabinet 3\\
\textcolor{case_purple}{\textbf{Observation 5:}} You close the cabinet 3.\\
. . .\\
\textcolor{red}{\textbf{Failure!}}

\begin{center}\textbf{Agent w/ IPR}\end{center}
\textcolor{case_blue}{\textbf{Thought 4:}} Cabinet 3 is closed. I'll open it to see if there's a pan inside.\\
\textcolor{case_blue}{\textbf{Action 4:}} open cabinet 3\\
\textcolor{case_purple}{\textbf{Observation 4:}} You open the cabinet 3. The cabinet 3 is open. In it, you see a plate 2.\\
\textcolor{case_blue}{\textbf{Thought 5:}} There's a plate 2 in cabinet 3, but no pan. I'll close it and check the next cabinet.\\
\textcolor{case_blue}{\textbf{Action 5:}} close cabinet 3\\
\textcolor{case_purple}{\textbf{Observation 5:}} You close the cabinet 3.\\
. . .\\
\textcolor{red}{\textbf{Failure!}}
\columnbreak

\begin{center}{\textbf{Agent w/ STeCa (Ours)}}\end{center}

\sethlcolor{green!50}{\textbf{\hl{Thought 4:}}} Cabinet 2 has a pan. I should take pan 1 from cabinet 2 instead of moving to cabinet 3. This will allow me to accomplish part of my task to find two pans. \textcolor{my_green}{\textit{\textbf{(\# Self-reflection)}}}\\
\sethlcolor{green!50}\textbf{\hl{Action 4:}} take pan 1 from cabinet 2\quad \\
\textcolor{case_purple}{\textbf{Observation 4:}} You pick up the pan 1 from cabinet 2.\\
\textcolor{case_blue}{\textbf{Thought 5:}} With pan 1 in hand, I need to place it on a countertop to keep it accessible while I search for the second pan.\\
\textcolor{case_blue}{\textbf{Action 5:}} go to countertop 1\\
\textcolor{case_purple}{\textbf{Observation 5:}} On the countertop 1, you see: a bread 1, a cup 1, a fork 2, a fork 1, a knife 3, . . .\\
\textcolor{case_blue}{\textbf{Thought 6:}} I am at countertop 1. I should place pan 1 here and ensure it is securely positioned so it doesn't fall off.\\
\textcolor{case_blue}{\textbf{Action 6:}} put pan 1 in/on countertop 1\quad \\
\textcolor{case_purple}{\textbf{Observation 6:}} You put the pan 1 in/on the countertop 1.\\
\textcolor{case_blue}{\textbf{Thought 7:}} Pan 1 is now on the countertop. I need to find the second pan. I will continue searching the cabinets.\\
\textcolor{case_blue}{\textbf{Action 7:}} go to cabinet 3\\
\textcolor{case_purple}{\textbf{Observation 7:}} On cabinet 3, you see: a pot 1 and a pan 2.\\
\textcolor{case_blue}{\textbf{Thought 8:}} In cabinet 3, I found another pan. I should take pan 2 to complete the task of finding two pans.\\
\textcolor{case_blue}{\textbf{Action 8:}} take pan 2 from cabinet 3\quad \\
\textcolor{case_purple}{\textbf{Observation 8:}} You pick up the pan 2 from cabinet 3.\\
\textcolor{case_blue}{\textbf{Thought 9:}} I have pan 2 now and need to place it on the same countertop as pan 1 to complete the task.\\
\textcolor{case_blue}{\textbf{Action 9:}} go to countertop 1\\
\textcolor{case_purple}{\textbf{Observation 9:}} On the countertop 1, you see: a bread 1, a cup 1, a fork 2, a fork 1, a knife 3, . . .\\
\textcolor{case_blue}{\textbf{Thought 10:}} I am back at countertop 1 with pan 2. I should place it next to pan 1, ensuring both are ready for use.\\
\textcolor{case_blue}{\textbf{Action 10:}} put pan 2 in/on countertop 1\quad \\
\textcolor{my_green}{\textbf{Success!}}
\end{multicols}
\end{tcolorbox}
\begin{figure}[ht]
    \centering
    \vspace{-8pt}
    \caption{
    Case study on the ALFWorld benchmark.
    }
    \label{fig:alfworld_case}
\end{figure}

\twocolumn

\section{Supplementary Experiments}

\subsection{Additional Evaluation on ScienceWorld}

To deeply validate our proposed method, we now additionally conducted experiments on the ScienceWorld~\citep{wang2022scienceworld} environment. It contains various long-horizon science experiment tasks, with an average steps of 14.4 in the collected trajectories. We compare our method with two repersentative baseline method: SFT and IPR. As shown in Table~\ref{tab:scienceworld_results}, the evaluation results reported below show that our method outperforms the compared mehtods. These reuslts consistently demonstrate the effectiveness of our method.

\begin{table}[ht]
\centering
\begin{tabular}{lcc}
\hline
\textbf{} & \textbf{Seen Tasks} & \textbf{Unseen Tasks} \\ \hline
SFT & 67.4 & 53.0 \\ 
IPR & 75.0 & 66.8 \\ 
\textbf{STeCa (Ours)} & \textbf{77.3} & \textbf{68.9} \\ \hline
\end{tabular}
\caption{Performance comparison of SFT, IPR, and STeCa on seen and unseen test-tasks in ScienceWorld.}
\label{tab:scienceworld_results}
\end{table}

\subsection{Affordance Analyses for LLM Agents}

\begin{figure}[ht]
  \includegraphics[width=1.0\linewidth]{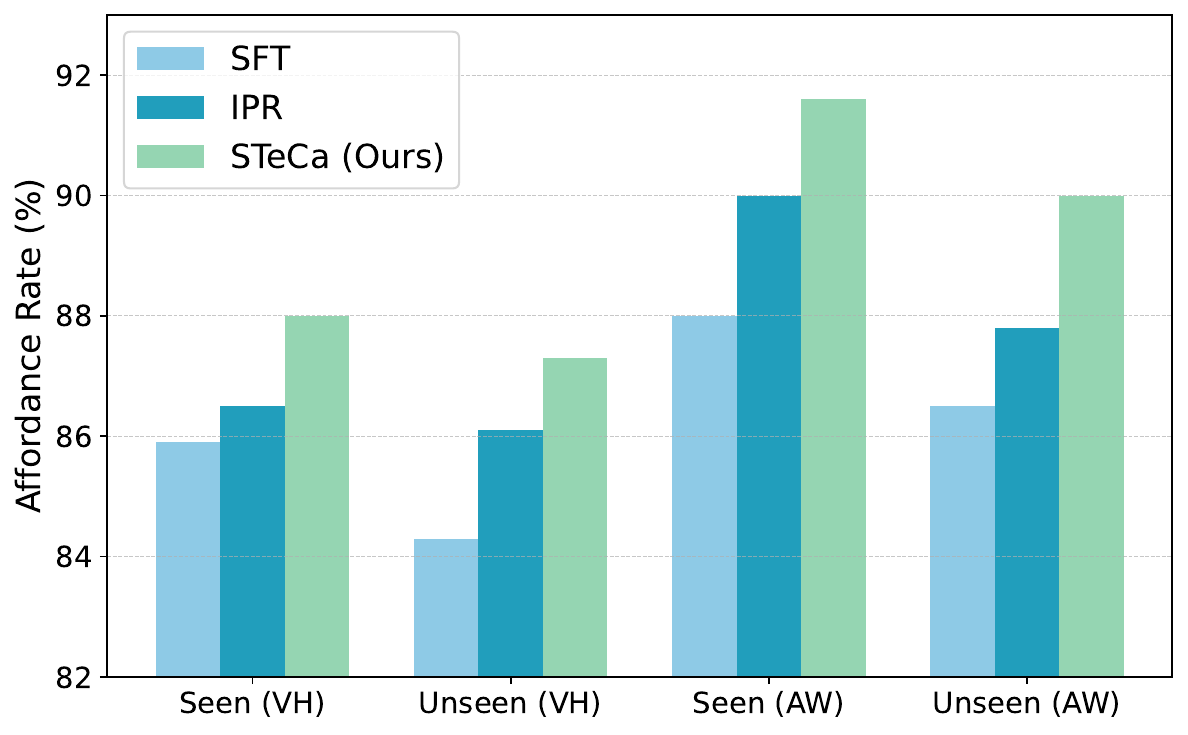}
  \caption{Comparison of affordance rates (\%) across different methods on seen and unseen tasks in VirtualHome (VH) and ALFWorld (AW).}
  \label{fig:affordance_rate}
\end{figure}

To make our assessments more comprehensive, we additionally utilized the affordance rate~\citep{wang-etal-2024-e2cl,ma2024agentboard} metric to evaluate the execution success rate of actions generated by LLM agents in VirtualHome and ALFWorld environments. The evaluation results are presented in Figure~\ref{fig:affordance_rate}. Our method consistently outperforms baseline methods in terms of affordance rate, generating more executable actions that satisfy environmental constraints.

\subsection{Ablation Studies}

We conducted an additional ablation study by individually removing different training losses, namely $L_{D_c}$ (calibration loss), $L_{D_e}$
 (exploration loss), and 
$L_{D_s}$ (success-guided loss), and reported the corresponding results in the Table~\ref{tab:loss_ablation}.
Our findings show that removing any of the loss terms results in a noticeable drop in performance, highlighting the complementary roles of the three components: 1) 
 helps the agent develop self-reflection capabilities, enabling it to adjust and calibrate its trajectories more effectively. 2) 
 enhances the agent’s ability to explore efficiently; removing it leads to reduced exploration effectiveness. 3) 
 improves the agent’s ability to complete more challenging tasks. These results underscore the importance of each loss term in contributing to the agent's overall performance.

\begin{table}[ht]
\centering
\begin{tabular}{lcc}
\hline
\textbf{} & \textbf{Unseen (VH)} & \textbf{Unseen (AW)} \\ \hline
\textbf{STeCa} & 63.6 & 76.1 \\ 
w/o $L_{D_c}$ & 62.2 & 75.1 \\ 
w/o $L_{D_e}$ & 60.5 & 71.2 \\ 
w/o $L_{D_b}$ & 61.5 & 73.9 \\ \hline
\end{tabular}
\caption{Ablation study on the success rate (\%) for unseen tasks in VirtualHome (VH) and ALFWorld (AW). The study evaluates the impact of individually removing different training losses: calibration loss ($L_{D_c}$), exploration loss ($L_{D_e}$), and success-guided loss ($L_{D_b}$).}
\label{tab:loss_ablation}
\end{table}

\subsection{Hyperparameter Analyses}

To analyze the impact of the hyperparameter $\delta$, we conducted additional experiments on the unseen test sets of ALFWorld and VirtualHome. The results, shown in Table~\ref{tab:hyperparam_delta}, demonstrate that the choice of $\delta$ significantly influences the model's performance. Specifically, setting $\delta = 0$ consistently yields the best results for both datasets, achieving a success rate of 76.1\% on ALFWorld and 63.6\% on VirtualHome.

Increasing $\delta$ slightly reduces performance, likely due to over-filtering valid calibration opportunities, which limits the agent's ability to identify and self-correct deviated actions. Conversely, setting $\delta$ to a negative value also degrades performance, as it may tolerate suboptimal actions excessively, leading to less precise deviation detection.

These findings highlight the importance of carefully tuning $\delta$ to balance between filtering suboptimal actions and retaining sufficient calibration opportunities. In our experiments, $\delta = 0$ emerges as the optimal setting, providing robust performance across both datasets.

\begin{table}[t]
\centering
\begin{tabular}{lcc}
\hline
   & \textbf{AW} & \textbf{VH} \\ \hline
$\delta = -0.01$ & 75.8 & 63.3 \\ 
$\delta = 0$ & \textbf{76.1} & \textbf{63.6} \\ 
$\delta = 0.05$ & 75.5 & 63.0 \\ 
$\delta = 0.1$ & 75.2 & 62.6 \\ \hline
\end{tabular}
\caption{Hyperparameter analyses of $\delta$ on the test sets of ALFWorld (AW) and VirtualHome (VH). The table reports the success rate (\%) across different values of $\delta$. The best results are highlighted in bold.}
\label{tab:hyperparam_delta}
\end{table}

\subsection{Discussions on Long-horizon Tasks}

Long-horizon tasks are a significant challenge in trajectory calibration, requiring agents to perform a relatively large number of sequential interaction steps. These tasks are especially prevalent in environments such as ALFWorld and VirtualHome, where tasks often involve complex sub-task sequences to achieve high-level objectives. For instance, in ALFWorld, tasks include subtasks such as placing clean potatoes into a microwave, cooling mugs, and arranging them into a coffee machine. Similarly, tasks in VirtualHome often require intricate action sequences, such as writing emails using a laptop or performing household cleaning.

To better characterize long-horizon tasks, we compare the average interaction steps required across multiple environments. As shown in Table~\ref{tab:avg_turns}, ALFWorld and VirtualHome demand significantly higher average steps (10.1 and 11.5, respectively) compared to other environments like WebShop, Maze, Wordle, and ToolQuery. These results highlight their suitability as benchmarks for long-horizon task evaluation.

\begin{table}[ht]
\centering
\resizebox{0.94\linewidth}{!}{
\begin{tabular}{lc}
\hline
\textbf{Environment} & \textbf{Avg. Turn} \\ \hline
WebShop~\citep{yao2022webshop}    & 5.1      \\
Maze~\citep{abdulhai2023lmrl}     & 4.3               \\
Wordle~\citep{abdulhai2023lmrl}    & 4.3               \\
ToolQuery~\citep{ma2024agentboard}   & 5.0               \\
ALFWorld~\citep{shridhar2020alfworld}   & 10.1         \\
VirtualHome~\citep{puig2018virtualhome}     & 11.5      \\ \hline
\end{tabular}
}
\caption{Average interaction steps required for tasks across different environments. ALFWorld and VirtualHome exhibit significantly longer task horizons compared to other environments.}
\label{tab:avg_turns}
\end{table}

To evaluate the effectiveness of our STeCa in addressing long-horizon tasks, we conducted experiments on the unseen set of VirtualHome tasks. Tasks were categorized into three groups based on the number of interaction steps required for completion: short-horizon tasks ($\leq7$ steps), medium-horizon tasks (7--13 steps), and long-horizon tasks ($>13$ steps). The results, shown in Table~\ref{tab:long_horizon_results}, demonstrate that STeCa consistently outperforms baseline methods (SFT and IPR) as task complexity increases. For short-horizon tasks, all methods perform similarly, with STeCa achieving a success rate of \textbf{76.2\%}, matching the best baseline (SFT). However, for medium-horizon tasks, STeCa outperforms the baselines, achieving \textbf{60.0\%}, compared to \textbf{50.5\%} (SFT) and \textbf{59.4\%} (IPR). The advantage becomes most evident for long-horizon tasks, where STeCa achieves \textbf{48.9\%}, compared to \textbf{40.0\%} (SFT) and \textbf{42.2\%} (IPR).

\begin{table}[t]
\centering
\begin{tabular}{lccc}
\hline
\textbf{Steps} & \textbf{SFT} & \textbf{IPR} & \textbf{STeCa (Ours)} \\ \hline
$\leq7$        & 76.2         & 74.3        & 76.2                 \\
7--13          & 50.5         & 59.4        & 60.0                 \\
$>13$          & 40.0         & 42.2        & \textbf{48.9}        \\ \hline
\end{tabular}
\caption{Success rate (\%) for tasks of varying lengths in the VirtualHome environment. STeCa consistently outperforms baselines, particularly in long-horizon tasks.}
\label{tab:long_horizon_results}
\end{table}



These findings confirm the robustness of STeCa in long-horizon scenarios. As the task horizon increases, the performance gap between STeCa and baseline methods widens, highlighting the increasing importance of an agent's calibration capability for maintaining effective trajectory adjustments in more complex and extended tasks. This demonstrates that STeCa is superior in addressing the challenges posed by long-horizon tasks.

\end{document}